\documentclass[conference]{IEEEtran}
\IEEEoverridecommandlockouts
\usepackage{cite}
\usepackage{amsmath,amssymb,amsfonts}
\usepackage{algorithmic}
\usepackage{graphicx}
\usepackage{textcomp}
\usepackage{xcolor}
\usepackage{graphicx}
\usepackage{caption}
\usepackage{subcaption}
\usepackage{multirow}
\usepackage{booktabs}

\def\BibTeX{{\rm B\kern-.05em{\sc i\kern-.025em b}\kern-.08em
    T\kern-.1667em\lower.7ex\hbox{E}\kern-.125emX}}
\begin{document}

\title{Robust and Generalizable Heart Rate Estimation via Deep Learning for Remote Photoplethysmography in Complex Scenarios\thanks{This work was supported in part by the National Natural Science Foundation of China under Grant 62471232, Grant 62401262, Grant 62301255, and Grant 62201259, in part by the Key Research and Development Plan of Jiangsu Province under Grant BE2023819, in part by the Key Program of the National Natural Science Foundation of China under Grant 62431013, and in part by the Joint Funds of the National Natural Science Foundation of China under Grant U24A20230.}}

\author{Kang Cen, Chang-Hong Fu*, Hong Hong \\
	\textit{School of Electronic and Optical Engineering} \\
	\textit{Nanjing University of Science and Technology}\\
	Nanjing, China \\
	\{ckang16671273780, enchfu, hongnju\}@\{foxmail.com, njust.edu.cn, njust.edu.cn\}
}

\maketitle

\begin{abstract}
Non-contact remote photoplethysmography (rPPG) technology enables heart rate measurement from facial videos. However, existing network models still face challenges in accuracy, robustness, and generalization capability under complex scenarios. This paper proposes an end-to-end rPPG extraction network that employs 3D convolutional neural networks to reconstruct accurate rPPG signals from raw facial videos. We introduce a differential frame fusion module that integrates differential frames with original frames, enabling frame-level representations to capture blood volume pulse (BVP) variations. Additionally, we incorporate Temporal Shift Module (TSM) with self-attention mechanisms, which effectively enhance rPPG features with minimal computational overhead. Furthermore, we propose a novel dynamic hybrid loss function that provides stronger supervision for the network, effectively mitigating overfitting. Comprehensive experiments were conducted on not only the PURE and UBFC-rPPG datasets but also the challenging MMPD dataset under complex scenarios, involving both intra-dataset and cross-dataset evaluations, which demonstrate the superior robustness and generalization capability of our network. Specifically, after training on PURE, our model achieved a mean absolute error (MAE) of 7.58 on the MMPD test set, outperforming the state-of-the-art models.
\end{abstract}


\section{Introduction}
Remote Photoplethysmography (rPPG) technology, as an emerging non-contact physiological signal monitoring method, eliminates the need for sensor attachment, avoiding discomfort and usage limitations compared to traditional contact-based methods like ECG and PPG.

Early rPPG methods primarily relied on signal processing approaches: blind source separation-based methods \cite{poh2010non,lewandowska2012measuring} and skin optical reflection model-based methods \cite{de2013robust,wang2016algorithmic}. However, rPPG signals exhibit low amplitude and are susceptible to noise, illumination variations, and motion artifacts, resulting in suboptimal performance in complex environments.

With deep learning advancement, data-driven approaches have gained prominence. Deep learning-based rPPG methods can be classified into non-end-to-end and end-to-end networks. Non-end-to-end networks \cite{niu2019rhythmnet,yang2018heart} require complex preprocessing and may introduce subjective bias. In contrast, end-to-end networks \cite{liu2020multi,chen2018deepphys,EfficientPhys,zou2025rhythmformer} integrate feature extraction, preprocessing, and classification into a unified model, minimizing human intervention.

Recent 2D convolution-based end-to-end networks include DeepPhys \cite{chen2018deepphys}, MTTS-CAN \cite{liu2020multi}, and EfficientPhys. However, 2D convolutional networks excel at spatial features but cannot directly capture temporal information, making it challenging to perceive dynamic changes between consecutive frames.

Transformer networks have garnered attention for their superior global information capture and long-term dependency handling capabilities. Nevertheless, Transformer architectures typically demand substantial computational resources due to complex mechanisms and extensive parameters.

To address these challenges, we propose physFSUNet based on 3D convolutional networks, which can simultaneously analyze spatial and temporal features while requiring fewer parameters than Transformer architectures. Our main contributions are:
\begin{itemize}
	\item Introducing temporal shift units into 3D convolutional networks for spatial information capture at minimal cost while reducing computational complexity;
	\item Proposing a dynamic learning hybrid loss function providing temporal and frequency domain supervision to mitigate overfitting;
	\item Incorporating fusion stem \cite{zou2025rhythmformer} into 3D convolutional networks to enhance frame-level perception capability;
	\item Demonstrating superior performance in both intra-dataset and cross-dataset experiments, including the challenging MMPD dataset \cite{tang2023mmpd}.
\end{itemize}

The rest of the paper is organized as follows. Section II introduces the methodology including the network model and proposed approach. Section III presents the experimental setup, datasets, and analyzes the experimental results. Section IV concludes the paper and discusses future work.

\section{METHODOLOGY}
Section A presents the overall framework of physFSUNet, followed by Section B which introduces the network's fusion structure—differential frame fusion. Section C introduce STAS Block,and Section D introduces the dynamic learning-based Hybrid Loss Function. 

\subsection{Network framework of physFSUNet}

\begin{figure*}[t]
	\centering
	\includegraphics[width=0.9\textwidth]{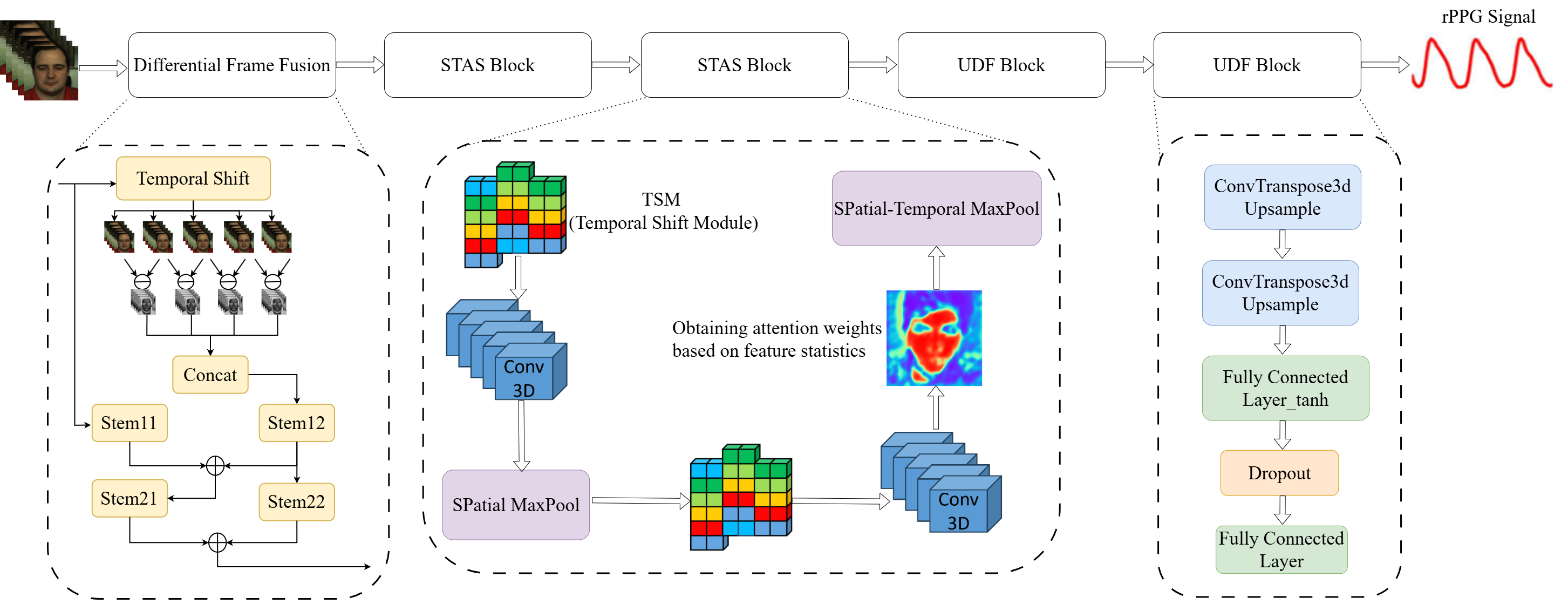}
	\caption{The physFSUNet architecture includes the Differential Frame Fusion module, Spatio-Temporal Attention Shift (STAS) Block, and Upsampling Decoding Fusion (UDF) Block.}
	\label{fig:framework}
\end{figure*}

As illustrated in Fig~\ref{fig:framework}, the physFSUNet architecture comprises three core components: the Differential Frame Fusion module, the Spatio-Temporal Attention Shift (STAS) Block, and the Upsampling Decoding Fusion (UDF) Block.

Initially, the network takes RGB video sequences as input $I_{\text{input}} \in \mathbb{R}^{3 \times D \times H \times W}$ and employs the Differential Frame Fusion module to integrate original frame information with differential frame information, enabling direct learning of frame-level rPPG information and reducing preprocessing complexity.

The features are then fed into the STAS Block for spatio-temporal physiological signal feature extraction. This module leverages the Temporal Shift Module (TSM)~\cite{Lin_Gan_Han_2019} to facilitate information exchange between adjacent frames and incorporates a self-attention mechanism to automatically locate and enhance pixels with strong physiological signals in skin regions.

Finally, the UDF Block processes the compressed features through upsampling layers and an MLP classifier to decode physiological parameters. The network outputs continuous rPPG waveform signals $S_{\text{rPPG}} \in \mathbb{R}^{1 \times T}$, achieving end-to-end mapping from RGB video to physiological signals.

\subsection{Differential Frame Fusion}
In remote photoplethysmography (rPPG) signal extraction, the selection of network input modalities represents a fundamental architectural decision that significantly impacts system performance. There are primarily two approaches for acquiring network inputs: using raw frames \cite{EfficientPhys,physnet,phsformer} and using normalized differential frames \cite{liu2020multi,chen2018deepphys}. 

Raw video frames preserve comprehensive spatiotemporal information but are vulnerable to photometric variations, environmental artifacts, and motion-induced disturbances. The high-dimensional nature introduces computational inefficiencies and impedes discrimination of subtle cardiovascular-related chromatic fluctuations from extraneous visual content.

Conversely, normalized differential frames employ temporal differentiation to enhance dynamic features and mitigate illumination variations, yet this preprocessing introduces systematic information degradation. The temporal derivative characteristics result in attenuation of static physiological baselines and can amplify noise propagation and motion artifacts.

To address these limitations, we propose a differential frame fusion module based on the Fusion Stem concept \cite{zou2025rhythmformer}. This approach combines raw frames with differential frames, leveraging the rich information content of raw frames to provide comprehensive contextual information while enabling the network to focus on frame-level transformations that are crucial for rPPG signal extraction. The differential frame module achieves enhanced rPPG feature extraction with minimal additional computational cost.

As illustrated in Fig~\ref{fig:framework}, The proposed architecture consists of two parallel branches. The right branch begins with a temporal shifting unit that generates time-shifted frames $D_{t-2}$, $D_{t-1}$, $D_t$, $D_{t+1}$, and $D_{t+2}$. Subsequently, differential operations are performed to obtain $D'_{-2}$, $D'_{-1}$, $D'_1$, and $D'_2$, which are then concatenated along the channel dimension. The resulting output has dimensions $I_{concat} \in \mathbb{R}^{12 \times D \times H \times W}$.

The concatenated differential frames are processed through the $\text{Stem}_{12}$ layer for initial fusion, producing $I_{diff}$ as described in Equation~(\ref{eq:diff}). The $\text{Stem}_{12}$ module consists of a 3D convolution with kernel size $5 \times 5$, stride 1, and padding 2, followed by batch normalization and ReLU activation.

\begin{equation}
	I_{diff} = \mathcal{F}_{\text{stem}_{12}}(\text{concat}(D'_{-2}, D'_{-1}, D'_1, D'_2))
	\label{eq:diff}
\end{equation}

The left branch directly processes raw frames through the $\text{Stem}_{11}$ module to obtain $I_{raw}$, as shown in Equation~(\ref{eq:raw}). The $\text{Stem}_{11}$ architecture is identical to $\text{Stem}_{12}$.

\begin{equation}
	I_{raw} = \mathcal{F}_{\text{stem}_{11}}(D_t)
	\label{eq:raw}
\end{equation}

Subsequently, the outputs $I_{raw}$ and $I_{diff}$ are fused using weighting coefficients $\alpha$ and $\beta$ to serve as input for the $\text{Stem}_{22}$ module. The $\text{Stem}_{22}$ module employs a 3D convolution with kernel size $3 \times 3$, stride 1, and padding 1, followed by batch normalization and ReLU activation. Concurrently, the left branch processes $I_{raw}$ through $\text{Stem}_{21}$ for feature enhancement. The final differential frame fusion $I_{stem}$ is obtained by combining the outputs according to the $\alpha$ and $\beta$ ratios, as expressed in Equation~(\ref{eq:fusion}). The $\text{Stem}_{21}$ architecture is identical to $\text{Stem}_{22}$.

\begin{IEEEeqnarray}{rCl}
	I_{stem} &=& \alpha \times \mathcal{F}_{\text{stem}_{21}}(I_{raw}) \nonumber\\
	&& + \beta \times \mathcal{F}_{\text{stem}_{22}}(\alpha \times I_{raw} + \beta \times I_{diff})
	\IEEEeqnarraynumspace
	\label{eq:fusion}
\end{IEEEeqnarray}

This differential frame fusion approach effectively combines the complementary advantages of both raw frames and differential frames while mitigating their respective limitations, resulting in improved rPPG signal extraction performance.

\subsection{STAS Block}

\begin{figure}[t]
	\centering
	\begin{subfigure}[t]{0.48\linewidth}
		\centering
		\includegraphics[width=\linewidth]{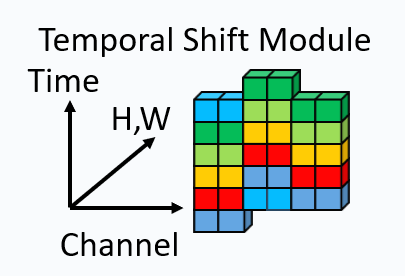}
		\caption{TSM}
		\label{fig:TSM}
	\end{subfigure}
	\hfill
	\begin{subfigure}[t]{0.35\linewidth}
		\centering
		\includegraphics[width=\linewidth]{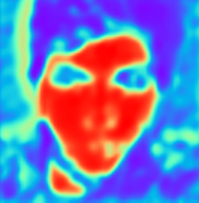}
		\caption{Self-attention mechanism}
		\label{fig:self_attention}
	\end{subfigure}
	\caption{The Temporal Shift Module (TSM) and the output feature maps of the self-attention mechanism. (a) TSM module; (b) Output feature map of the self-attention mechanism.}
	\label{fig:both_datasets}
\end{figure}

The STAS Block comprises four core components: TSM-based three-dimensional convolutional layers, self-attention mechanisms, maximum spatial pooling layers, and maximum spatiotemporal pooling layers. 

Input information is first processed through TSM-based three-dimensional convolutional layers, which can more efficiently capture temporal dynamic information compared to conventional three-dimensional convolutions. As illustrated in Fig~\ref{fig:TSM}, the TSM module equally divides the input tensor into three blocks along the channel dimension, then shifts the first block forward by one frame, shifts the second block backward by one frame, while the third block remains unchanged in the temporal dimension. This provides a lightweight solution that effectively integrates temporal information while significantly reducing computational overhead.

Subsequently, a self-attention mechanism is applied to optimize the features. The feature map after optimization is shown in Fig~\ref{fig:self_attention}. Considering the presence of multiple noise sources and that temporal shifting operations may introduce additional noise, we introduce a soft attention mechanism to focus on pixels containing physiological signals. The attention mask assigns higher weights to skin regions with stronger signal intensity through multi-channel fusion followed by a softmax attention layer with one-dimensional convolution and sigmoid activation. The computation method for the attention mask is shown in Equation~(\ref{eq:attention_mask}).

The Equation~(\ref{eq:attention_mask}) proposed in \cite{EfficientPhys} is defined as:
\begin{equation}
	\left( w_{c}^{i} \cdot \text{ts}\left( \mathbb{Z}_{a}^{i} \right) + b_{c}^{i} \right) \odot \frac{H_i W_i \cdot \sigma \left( w_{a}^{i} \mathbb{Z}_{a}^{i} + b_{a}^{i} \right)}{2\left\| \sigma \left( w_{a}^{i} \mathbb{Z}_{a}^{i} + b_{a}^{i} \right) \right\|_1}
	\label{eq:attention_mask}
\end{equation}

where $i$ denotes the layer index, $w_a^i$ represents the $1 \times 1$ convolution kernel for self-attention in the $i$-th layer, followed by a sigmoid activation function $\sigma(\cdot)$. L1 normalization is applied to soften extreme values in the mask, ensuring the network avoids pixel anomalies. $w_c^i$ denotes the two-dimensional convolution kernel in the $i$-th layer, followed by a tensor shift module. $\text{ts}(\cdot)$ represents the tensor shift operation, $\odot$ denotes element-wise multiplication, $\mathbb{Z}_a^i$ represents the current input tensor, and $H_i, W_i$ represent the height and width dimensions, respectively.

To achieve more refined feature extraction, we design a progressive feature refinement mechanism based on three core considerations: multi-scale feature fusion for capturing long-range dependencies within broader spatiotemporal ranges, hierarchical feature representation from low-level temporal patterns to high-level semantic features, and progressive noise suppression through two-level processing.

After optimization through the self-attention mechanism, feature information is processed through the maximum spatiotemporal pooling layer. Spatiotemporal pooling performs feature aggregation simultaneously across temporal and spatial dimensions, effectively preserving temporal continuity while achieving spatial invariance. Additionally, the network introduces Dropout regularization to enhance generalization capability and prevent overfitting. This dual TSM-attention architecture combined with hierarchical pooling strategies enables the STAS Block to achieve precise modeling of complex spatiotemporal dynamic patterns while maintaining computational efficiency.

\subsection{Dynamic Learning-Based Hybrid Loss Function}
Temporal-based objectives capture signal morphology and trends with computational efficiency but may inadequately address the periodic nature of cardiovascular signals. Conversely, frequency-domain objectives enforce spectral consistency yet struggle with noise characteristics in realistic physiological measurements.

A critical limitation emerges from the mismatch between blood volume pulse (BVP) reconstruction accuracy and clinically relevant heart rate (HR) estimation performance. To address this, we introduce a probabilistic HR-based loss function \cite{zou2025rhythmformer} that models HR as a stochastic variable following a normal distribution, acknowledging inherent uncertainty in ground truth acquisition. The heart rate distribution is characterized as equation~(\ref{eq:hr_distribution}):

\begin{equation}
	HR_x = N(\mu, \sigma^2)
	\label{eq:hr_distribution}
\end{equation}

where:
\begin{itemize}
	\item $N$ represents the normal distribution
	\item $x$ represents the predicted or actual signal
	\item $\mu$ represents the mean of the normal distribution, specifically:
\end{itemize}

\begin{equation}
	\mu = \arg\max_f(\text{PSD}(BVP_x))
\end{equation}

where $\arg\max_f$ identifies the frequency $f$ corresponding to the maximum spectral power, and PSD denotes the power spectral density transformation.

The distribution variance parameter $\sigma$ is empirically configured to 3.0 based on physiological heart rate variability characteristics.

Therefore, the HR distance loss can be expressed as equation~(\ref{eq:hr_loss}), where subscript gt denotes ground truth and subscript pred denotes predicted values. KL represents KL divergence.

\begin{equation}
	\mathcal{L}_{HR} = \text{KL}(HR_{gt}, HR_{pred})
	\label{eq:hr_loss}
\end{equation}

Our temporal constraint $\mathcal{L}_{\text{Time}}$ leverages the negative Pearson correlation coefficient to capture signal coherence, as formulated in equation~(\ref{eq:time_loss}). The spectral constraint $\mathcal{L}_{\text{Freq}}$ employs cross-entropy between the power spectral density distributions of predicted and reference BVP signals at their respective dominant frequency components, as detailed in equation~(\ref{eq:freq_loss}). Drawing inspiration from progressive learning strategies \cite{phsformer}, we implement an adaptive weighting scheme that gradually modulates the influence of frequency-domain constraints throughout training, thereby mitigating potential overfitting issues. The composite loss function is formulated in equation~(\ref{eq:total_loss}).

\begin{equation}
	\mathcal{L}_{\text{Time}} = 1 - \frac{\sum_{i=1}^{n}(X_i - \bar{X})(Y_i - \bar{Y})}{\sqrt{\sum_{i=1}^{n}(X_i - \bar{X})^2} \cdot \sqrt{\sum_{i=1}^{n}(Y_i - \bar{Y})^2}}
	\label{eq:time_loss}
\end{equation}

\begin{equation}
	\mathcal{L}_{\text{Freq}} = \text{CE}(\max \text{Idx}(\text{PSD}(BVP_{gt})), \text{PSD}(BVP_{pred}))
	\label{eq:freq_loss}
\end{equation}

\begin{equation}
	\mathcal{L}_{\text{overall}} = \alpha \cdot \mathcal{L}_{\text{Time}} + \beta \cdot (\mathcal{L}_{CE} + \mathcal{L}_{HR})
	\label{eq:total_loss}
\end{equation}

\begin{equation}
	\beta = \lambda \cdot \theta^{\frac{(\text{Epoch}_{\text{current}} - 1)}{\text{Epoch}_{\text{total}}}}
	\label{eq:beta}
\end{equation}

Empirically, we configure the optimization parameters as $\alpha = \beta = \lambda = 1.0$ and $\theta = 1.5$ for balanced multi-objective learning.
	
\section{Experiment}
\subsection{Dataset and performance metric}
This study employs three publicly accessible datasets for validation: PURE~\cite{pure}, UBFC-rPPG~\cite{ubfc}, and MMPD~\cite{tang2023mmpd} datasets. The PURE dataset comprises facial video recordings of 10 subjects under six different motion conditions, while the UBFC-rPPG dataset captures facial videos of 42 subjects in controlled indoor environments. Both PURE and UBFC-rPPG represent relatively simple scenarios with limited variations in lighting and environmental conditions.

In contrast, the MMPD dataset serves as a significantly more challenging benchmark, encompassing 33 diverse subjects with Fitzpatrick skin types 3-6 performing four typical activities (static, head rotation, verbal communication, and walking) under four distinct lighting conditions (high LED, low LED, incandescent, and natural light). The MMPD dataset contains 660 one-minute video segments with resolution of $320 \times 240$ pixels and 30Hz sampling rate. Its complex multimodal experimental design with diverse skin tones, varied lighting conditions, and dynamic activities makes it substantially more difficult than conventional datasets, providing a rigorous evaluation platform for algorithm robustness. The experimental evaluation employs four standard metrics: Mean Absolute Error (MAE), Root Mean Square Error (RMSE), Mean Absolute Percentage Error (MAPE), and Pearson correlation coefficient ($\rho$).
Most data for the comparison methods in the table come from original paper results and existing related studies. However, the results for training on MMPD and testing on UBFC-rPPG are our own experimental results. The data for methods that were not experimentally tested on specific datasets are marked with ''-''.
The best results are displayed in \textbf{bold}, and the second-best results are displayed in \textbf{\underline{bold with underline}}.

\subsection{Implementation details}
All experiments are conducted using the PyTorch-based open-source rPPG toolbox~\cite{rppg_toolbox}. Face detection is performed using the Haar Cascade algorithm in preprocessing. Training employs the AdamW optimizer with initial learning rate of $9 \times 10^{-3}$ and zero weight decay. When training on MMPD dataset, we strategically select a subset including LIGHT(1,2,3), MOTION(1), EXERCISE(2), and SKIN\_COLOR(3), as the target test dataset UBFC-rPPG represents simpler conditions and using the complete MMPD dataset would introduce excessive noise. Post-processing applies second-order Butterworth filtering and Welch algorithm for power spectral density computation. Batch size is set to 4 with 30 training epochs on NVIDIA RTX A5000 GPU.

\subsection{Intra-Dataset Evaluation}

\begin{table}[t]
	\renewcommand{\arraystretch}{1.3}
	\caption{Inner-dataset results on PURE, UBFC-rPPG, and MMPD (entire dataset)}
	\label{tab:inner_dataset_results}
	\centering
	\begin{tabular}{|l|l|c|c|c|c|}
		\hline
		\textbf{Method} & \textbf{Dataset} & \textbf{MAE$\downarrow$} & \textbf{RMSE$\downarrow$} & \textbf{MAPE$\downarrow$} & \textbf{$r\uparrow$} \\
		\hline
		\multirow{3}{*}{PhysNet~\cite{physnet}} & PURE & 2.10 & 2.60 & - & 0.99 \\
		& UBFC & 2.95 & 3.67 & - & 0.97 \\
		& MMPD & 4.80 & 11.80 & 4.74 & 0.60 \\
		\hline
		\multirow{3}{*}{TS-CAN~\cite{liu2020multi}} & PURE & 2.48 & 9.01 & - & 0.92 \\
		& UBFC & 1.70 & 2.72 & - & 0.99 \\
		& MMPD & 9.71 & 17.22 & 10.36 & 0.44 \\
		\hline
		\multirow{3}{*}{PhysFormer~\cite{phsformer}} & PURE & \textbf{\underline{1.10}} & \textbf{\underline{1.75}} & - & \textbf{0.99} \\
		& UBFC & \textbf{0.50} & \textbf{0.71} & - & \textbf{0.99} \\
		& MMPD & 11.99 & 18.41 & 12.50 & 0.18 \\
		\hline
		\multirow{2}{*}{EfficientPhys~\cite{EfficientPhys}} & PURE & - & - & - & - \\
		&UBFC & 1.14 & 1.81 & - & 0.99 \\
		& MMPD & 13.47 & 21.32 & 14.19 & 0.21 \\
		\hline
		\multirow{3}{*}{RhythmFormer~\cite{zou2025rhythmformer}} & PURE & \textbf{0.27} & \textbf{0.47} & - & \textbf{0.99} \\
		& UBFC & \textbf{0.50} & \textbf{\underline{0.78}} & - & \textbf{0.99} \\
		& MMPD & \textbf{3.07} & \textbf{6.81} & \textbf{3.24} & \textbf{0.86} \\
		\hline
		\multirow{3}{*}{Ours} & PURE & 2.05 & 9.15 & 2.65 & 0.96 \\
		& UBFC & \textbf{\underline{2.56}} & 4.29 & 2.85 & 0.98 \\
		& MMPD & \textbf{\underline{3.67}} & \textbf{\underline{9.39}} & \textbf{\underline{4.42}} & \textbf{\underline{0.79}} \\
		\hline
	\end{tabular}
\end{table}

For evaluation, we followed established protocols: PURE dataset used 60\% for training and 40\% for testing \cite{pure_ubfc_eval27, pure_eva39}, while UBFC dataset used the first 30 samples for training and last 12 for testing \cite{pure_ubfc_eval27, ubfc_eva38}. As shown in Table~\ref{tab:inner_dataset_results}, all models achieved strong performance on both datasets due to their relatively low complexity.

To assess generalization ability in complex scenarios, we evaluated on the challenging MMPD dataset with a 7:1:2 train/validation/test split \cite{zou2025rhythmformer}. Our method demonstrated strong competitiveness, achieving MAE of 3.67, RMSE of 9.39, MAPE of 4.42, and Pearson correlation coefficient ($\rho$) of 0.79, ranking second among state-of-the-art approaches.Overall, physFSUNet leverages 3D-CNN and STAS module for fine-grained rPPG modeling, demonstrating strong robustness for real-world applications.

\subsection{Cross-dataset evaluation}
	\begin{table}[t]
		\renewcommand{\arraystretch}{1.3}
		\caption{Cross-dataset results: training on PURE and UBFC-rPPG, testing on MMPD (entire dataset)}
		\label{tab:mmpdtesting}
		\centering
		\begin{tabular}{|l|l|c|c|c|c|}
			\hline
			\textbf{Method} & \textbf{Train} & \textbf{MAE$\downarrow$} & \textbf{RMSE$\downarrow$} & \textbf{MAPE$\downarrow$} & \textbf{$\rho\uparrow$} \\
			\hline
			\multirow{3}{*}{PhysNet~\cite{physnet}} & PURE & 13.94 & 21.61 & 15.15 & 0.20 \\
			& UBFC & 9.47 & 16.01 & 11.11 & 0.31 \\
			\hline
			\multirow{2}{*}{TS-CAN~\cite{liu2020multi}} & PURE & 13.94 & 21.61 & 15.15 & 0.20 \\
			& UBFC & 14.01 & 21.04 & 15.48 & 0.24 \\
			\hline
			\multirow{2}{*}{PhysFormer~\cite{phsformer}} & PURE & 10.49 & 17.14 & 11.98 & 0.32 \\
			& UBFC & 10.69 & 17.23 & 12.48 & 0.27 \\
			\hline
			\multirow{2}{*}{EfficientPhys~\cite{EfficientPhys}} & PURE & 14.03 & 21.62 & 15.32 & 0.17 \\
			& UBFC & 13.78 & 22.25 & 15.15 & 0.09 \\
			\hline
			\multirow{2}{*}{RhythmFormer~\cite{zou2025rhythmformer}} & PURE & \textbf{\underline{8.98}} & \textbf{\underline{14.85}} & \textbf{\underline{11.11}} & \textbf{0.51} \\
			& UBFC & \textbf{\underline{9.08}} & \textbf{\underline{15.07}} & \textbf{\underline{11.17}} & \textbf{0.53} \\
			\hline
			\multirow{2}{*}{Ours} & PURE & \textbf{7.58} & \textbf{14.12} & \textbf{8.88} & \textbf{\underline{0.50}} \\
			& UBFC & \textbf{8.54} & \textbf{14.27} & \textbf{10.81} & \textbf{\underline{0.47}} \\
			\hline
		\end{tabular}
	\end{table}
	
	\begin{table}[t]
		\renewcommand{\arraystretch}{1.3}
		\caption{Cross-dataset results: training on PURE and MMPD (subset), testing on UBFC-rPPG}
		\label{tab:ubfctesing}
		\centering
		\begin{tabular}{|l|l|c|c|c|c|}
			\hline
			\textbf{Method} & \textbf{Train} & \textbf{MAE$\downarrow$} & \textbf{RMSE$\downarrow$} & \textbf{MAPE$\downarrow$} & \textbf{$\rho\uparrow$} \\
			\hline
			\multirow{2}{*}{PhysNet~\cite{physnet}} & PURE & 1.63 & 3.78 & 1.68 & 0.97 \\
			& MMPD & \textbf{\underline{8.97}} & \textbf{\underline{17.25}} & \textbf{\underline{8.25}} & \textbf{\underline{0.61}} \\
			\hline
			\multirow{2}{*}{TS-CAN~\cite{liu2020multi}} & PURE &\textbf{\underline{1.29}} & \textbf{\underline{2.87}} & \textbf{\underline{1.50}} & \textbf{0.98} \\
			& MMPD & 9.06 & 20.46 & 8.16 & 0.50 \\
			\hline
			\multirow{2}{*}{PhysFormer~\cite{phsformer}} & PURE & 1.44 & 3.76 & 1.66 & 0.97 \\
			& MMPD & 20.80 & 29.05 & 18.99 & 0.08 \\
			\hline
			\multirow{2}{*}{EfficientPhys~\cite{EfficientPhys}} & PURE& 2.07 & 6.32 & 2.09 & 0.93 \\
			& MMPD & 17.15 & 27.33 & 16.29 & 0.44 \\
			\hline
			\multirow{2}{*}{RhythmFormer~\cite{zou2025rhythmformer}} & PURE & 1.61 & 3.78 & 1.73 & 0.97 \\
			& MMPD & 23.64 & 32.17 & 21.26 & 0.11 \\
			\hline
			\multirow{2}{*}{Ours} & PURE & \textbf{1.23} & \textbf{2.78} & \textbf{1.36} & \textbf{0.98} \\
			& MMPD & \textbf{5.71} & \textbf{13.92} & \textbf{5.05} & \textbf{0.70} \\
			\hline
		\end{tabular}
	\end{table}
	
Additionally, we follow the cross-dataset evaluation protocol outlined in the rPPG toolbox~\cite{rppg_toolbox} . Models are trained on the PURE and UBFC-rPPG datasets, as well as a subset of the MMPD dataset, with an 80\%-20\% train-validation split. Tables~\ref{tab:mmpdtesting}and ~\ref{tab:ubfctesing} present comparison results with state-of-the-art end-to-end methods.
Table~\ref{tab:mmpdtesting} shows the results when training on PURE and UBFC-rPPG datasets and testing on the MMPD dataset. Our network achieves first place in three out of four metrics, obtaining the lowest MAE, RMSE, and MAPE, while achieving the second-best Pearson correlation coefficient. Table~\ref{tab:ubfctesing} presents the results when training on PURE and MMPD (subset) datasets and testing on the UBFC-rPPG dataset. Our network ranks first across four metrics (MAE,RMSE,MAPE, and correlation) with significant margins.

\subsection{Ablation study}
\begin{table}[!t]
	\caption{Differential frame fusion ablation experiment}
	\label{tab:fusion_stem_ablation}
	\centering
	\footnotesize
	\begin{tabular}{|l|c|c|c|c|}
		\hline
		\textbf{Input} & \textbf{MAE$\downarrow$} & \textbf{RMSE$\downarrow$} & \textbf{MAPE$\downarrow$} & \textbf{$\rho\uparrow$} \\
		\hline
		Raw frame& 18.20 & 22.42 & 24.13 & 0.00 \\
		\hline
		Diff frame & 11.62 & 18.93 & 12.81 & 0.25 \\
		\hline
		Differential frame fusion & \textbf{7.58} & \textbf{14.12} & \textbf{8.88} & \textbf{0.50} \\
		\hline
	\end{tabular}
\end{table}

\begin{table}[!t]
	\caption{TSM module ablation experiment}
	\label{tab:tsm_module_ablation}
	\centering
	\footnotesize
	\begin{tabular}{|l|c|c|c|c|}
		\hline
		\textbf{Input} & \textbf{MAE$\downarrow$} & \textbf{RMSE$\downarrow$} & \textbf{MAPE$\downarrow$} & \textbf{$\rho\uparrow$} \\
		\hline
		w/o & 8.31 & 14.77 & 9.73 & 0.45 \\
		\hline
		w. & \textbf{7.58} & \textbf{14.12} & \textbf{8.88} & \textbf{0.50} \\
		\hline
	\end{tabular}
\end{table}

\begin{table}[!t]
	\caption{Self-attention mechanism ablation experiment}
	\label{tab:self_attention_ablation}
	\centering
	\footnotesize
	\begin{tabular}{|l|c|c|c|c|}
		\hline
		\textbf{Input} & \textbf{MAE$\downarrow$} & \textbf{RMSE$\downarrow$} & \textbf{MAPE$\downarrow$} & \textbf{$\rho\uparrow$} \\
		\hline
		w/o & 9.30 & 15.44 & 10.61 & 0.42 \\
		\hline
		w. & \textbf{7.58} & \textbf{14.12} & \textbf{8.88} & \textbf{0.50} \\
		\hline
	\end{tabular}
\end{table}

\begin{table}[!t] 
	\caption{Loss function ablation experiment}
	\label{tab:loss_function_ablation} 
	\centering 
	\footnotesize 
	\begin{tabular}{|l|c|c|c|c|} 
		\hline 
		\textbf{Loss Function} & \textbf{MAE$\downarrow$} & \textbf{RMSE$\downarrow$} & \textbf{MAPE$\downarrow$} & \textbf{$\rho\uparrow$} \\ 
		\hline 
		MSE & 9.36 & 15.81 & 10.69 & 0.37 \\ 
		\hline 
		Neg\_Pearson & 8.76 & 14.94 & 10.15 & 0.43 \\ 
		\hline 
		Neg\_Pearson+CE & 8.40 & 14.57 & 9.62 & 0.46 \\ 
		\hline 
		Neg\_Pearson+CE+KL & 8.40 & 14.57 & 9.62 & 0.46 \\ 
		\hline 
		Dynamic\_Neg\_Pearson+CE & 8.83 & 15.34 & 10.33 & 0.40 \\ 
		\hline 
		Dynamic learning-based hybrid & \textbf{7.58} & \textbf{14.12} & \textbf{8.88} & \textbf{0.50} \\ 
		\hline 
	\end{tabular} 
\end{table}
We conducted ablation experiments by training on PURE dataset and testing on MMPD dataset to evaluate the impact of different modules.

\textbf{ Impact of Fusion Stem}
As shown in Table~\ref{tab:fusion_stem_ablation}, the differential fusion stem achieved nearly 11 percentage points MAE improvement over raw frame input and 4 percentage points over differential frame input. By integrating differential frame information into raw frames, the differential fusion stem enables frame-level BVP waveform perception, effectively enhancing rPPG signal extraction while simplifying preprocessing.

\textbf{ Impact of TSM Module}
Table~\ref{tab:tsm_module_ablation} demonstrates that the TSM module facilitates rPPG signal extraction, achieving nearly 1 percentage point MAE improvement through simple temporal shift operations with minimal additional computation.

\textbf{ Impact of Self-Attention Mechanism}
As shown in Table~\ref{tab:self_attention_ablation}, the self-attention mechanism achieved nearly 2 percentage points MAE improvement by helping the network focus on pixels with higher physiological signal intensity and reducing noise introduced by TSM operations.

\textbf{ Impact of Loss Function}
Table~\ref{tab:loss_function_ablation} shows that our hybrid loss function with dynamic learning achieved 1-2 percentage points MAE improvement compared to simple MSE or negative Pearson loss functions. The dynamic weighting strategy uses smaller frequency domain weights in early training for rapid learning, then increases frequency domain weights in later stages to learn periodic characteristics and enhance generalization capability, preventing overfitting.

\subsection{Computational cost}
\begin{table}[t]
	\renewcommand{\arraystretch}{1.3}
	\caption{Model performance and computational cost comparison}
	\label{tab:model_comparison}
	\centering
	\begin{tabular}{|l|c|c|c|}
		\hline
		\textbf{Method} & \textbf{Parameters} & \textbf{MAE$\downarrow$} & \textbf{Memory (MB)} \\
		\hline
		DeepPhys~\cite{chen2018deepphys} & 7.504M & 16.92 & 4724.68 \\
		\hline
		EfficientPhys~\cite{EfficientPhys} & 7.439M & 14.03 & 4695.23 \\
		\hline
		TS-CAN~\cite{liu2020multi}& 7.504M & 13.94 & 4925.63 \\
		\hline
		PhysNet~\cite{physnet} & \textbf{769K} & 13.94 & 4978.92 \\
		\hline
		PhysFormer~\cite{phsformer} & 7.381M & 10.49 & 1012.51 \\
		\hline
		RhythmFormer~\cite{zou2025rhythmformer} & 3.251M & 9.73 & 1051.06 \\
		\hline
			Ours & \textbf{\underline{831K}} & \textbf{7.58} & \textbf{3375.00} \\
		\hline
	\end{tabular}
\end{table}
\begin{figure}[t]
	\centering
	\includegraphics[width=\columnwidth]{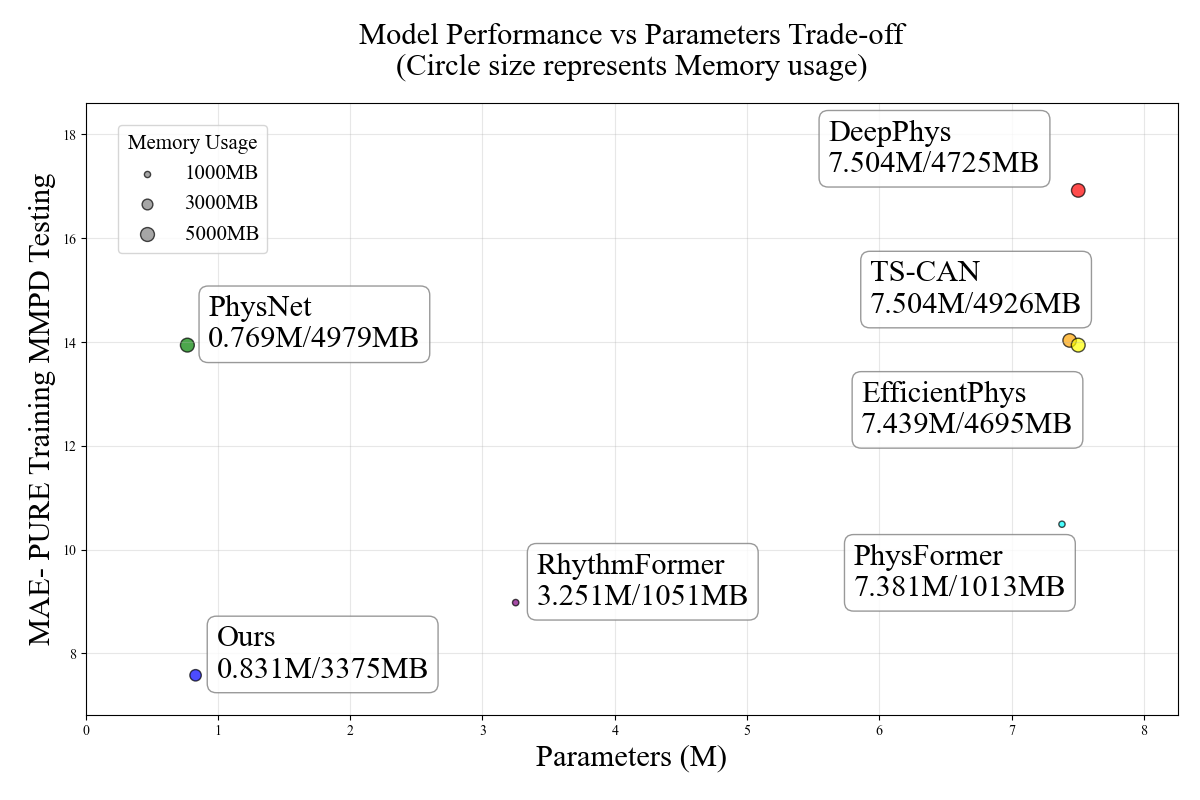}
	\caption{The horizontal axis is the parameter quantity of the model, and the vertical axis is the MAE value}
	\label{fig:params compare}
\end{figure}
Based on Fig~\ref{fig:params compare} and Table~\ref{tab:model_comparison}, our proposed network demonstrates significant technical advantages. From parameter efficiency perspective, our method achieves the best MAE performance of 7.58 using only 831K parameters, representing an 80-90\% reduction compared to other methods (3.25M-7.5M parameters). While substantially reducing model complexity, our method outperforms the second-best RhythmFormer (8.98 MAE) by 15.6\% and achieves 45.6\% improvement over PhysNet with similar parameter count (769K parameters, 13.94 MAE).

From computational resource perspective, our method exhibits reasonable memory management with 3375MB usage, demonstrating superior efficiency compared to parameter-heavy methods like DeepPhys and EfficientPhys (over 4500MB). Although memory usage is slightly higher than PhysFormer and RhythmFormer (~1000MB), this trade-off is justified by significant performance gains.

Fig~\ref{fig:params compare} illustrates our approach occupying the ideal lower-left region, representing the Pareto-optimal "low parameters-low error" solution. Table~\ref{tab:model_comparison} quantitatively validates our method's capability to achieve optimal performance with minimal parameters. These results demonstrate that our method successfully addresses the fundamental trade-off between model complexity and performance, providing a novel pathway for lightweight high-performance heart rate detection models.

\section{CONCLUSION}

This study proposes the physFSUNet network driven by a dynamic learning hybrid loss function, which integrates a 3D differential frame fusion module and STAS Block. Through intra-dataset experiments and cross-dataset validation, the network demonstrates superior performance in complex scenarios with excellent generalization capability. Ablation experiments confirm the positive contributions of each core module to physiological signal extraction, and these modules exhibit plug-and-play characteristics that can effectively enhance the rPPG signal extraction accuracy of other networks. Future research will focus on further optimizing the network's real-time performance.

\bibliographystyle{IEEEtran}  
\bibliography{main}

\begin{thebibliography}{10}
\providecommand{\url}[1]{#1}
\csname url@samestyle\endcsname
\providecommand{\newblock}{\relax}
\providecommand{\bibinfo}[2]{#2}
\providecommand{\BIBentrySTDinterwordspacing}{\spaceskip=0pt\relax}
\providecommand{\BIBentryALTinterwordstretchfactor}{4}
\providecommand{\BIBentryALTinterwordspacing}{\spaceskip=\fontdimen2\font plus
\BIBentryALTinterwordstretchfactor\fontdimen3\font minus
  \fontdimen4\font\relax}
\providecommand{\BIBforeignlanguage}[2]{{%
\expandafter\ifx\csname l@#1\endcsname\relax
\typeout{** WARNING: IEEEtran.bst: No hyphenation pattern has been}%
\typeout{** loaded for the language `#1'. Using the pattern for}%
\typeout{** the default language instead.}%
\else
\language=\csname l@#1\endcsname
\fi
#2}}
\providecommand{\BIBdecl}{\relax}
\BIBdecl

\bibitem{poh2010non}
M.-Z. Poh, D.~J. McDuff, and R.~W. Picard, ``Non-contact, automated cardiac
  pulse measurements using video imaging and blind source separation.''
  \emph{Optics express}, vol.~18, no.~10, pp. 10\,762--10\,774, 2010.

\bibitem{lewandowska2012measuring}
M.~Lewandowska and J.~Nowak, ``Measuring pulse rate with a webcam,''
  \emph{Journal of Medical Imaging and Health Informatics}, vol.~2, no.~1, pp.
  87--92, 2012.

\bibitem{de2013robust}
G.~De~Haan and V.~Jeanne, ``Robust pulse rate from chrominance-based rppg,''
  \emph{IEEE transactions on biomedical engineering}, vol.~60, no.~10, pp.
  2878--2886, 2013.

\bibitem{wang2016algorithmic}
W.~Wang, A.~C. Den~Brinker, S.~Stuijk, and G.~De~Haan, ``Algorithmic principles
  of remote ppg,'' \emph{IEEE Transactions on Biomedical Engineering}, vol.~64,
  no.~7, pp. 1479--1491, 2016.

\bibitem{niu2019rhythmnet}
X.~Niu, S.~Shan, H.~Han, and X.~Chen, ``Rhythmnet: End-to-end heart rate
  estimation from face via spatial-temporal representation,'' \emph{IEEE
  Transactions on Image Processing}, vol.~29, pp. 2409--2423, 2019.

\bibitem{yang2018heart}
W.~Yang, X.~Li, and B.~Zhang, ``Heart rate estimation from facial videos based
  on convolutional neural network,'' in \emph{2018 International Conference on
  Network Infrastructure and Digital Content (IC-NIDC)}.\hskip 1em plus 0.5em
  minus 0.4em\relax IEEE, 2018, pp. 45--49.

\bibitem{liu2020multi}
X.~Liu, J.~Fromm, S.~Patel, and D.~McDuff, ``Multi-task temporal shift
  attention networks for on-device contactless vitals measurement,''
  \emph{Advances in Neural Information Processing Systems}, vol.~33, pp.
  19\,400--19\,411, 2020.

\bibitem{chen2018deepphys}
W.~Chen and D.~McDuff, ``Deepphys: Video-based physiological measurement using
  convolutional attention networks,'' in \emph{Proceedings of the european
  conference on computer vision (ECCV)}, 2018, pp. 349--365.

\bibitem{EfficientPhys}
X.~Liu, B.~Hill, Z.~Jiang, S.~Patel, and D.~McDuff, ``Efficientphys: Enabling
  simple, fast and accurate camera-based cardiac measurement,'' in \emph{2023
  IEEE/CVF Winter Conference on Applications of Computer Vision (WACV)}, 2023,
  pp. 4997--5006.

\bibitem{zou2025rhythmformer}
B.~Zou, Z.~Guo, J.~Chen, J.~Zhuo, W.~Huang, and H.~Ma, ``Rhythmformer:
  Extracting patterned rppg signals based on periodic sparse attention,''
  \emph{Pattern Recognition}, vol. 164, p. 111511, 2025.

\bibitem{tang2023mmpd}
J.~Tang, K.~Chen, Y.~Wang, Y.~Shi, S.~Patel, D.~McDuff, and X.~Liu, ``Mmpd:
  Multi-domain mobile video physiology dataset,'' in \emph{2023 45th Annual
  International Conference of the IEEE Engineering in Medicine \& Biology
  Society (EMBC)}.\hskip 1em plus 0.5em minus 0.4em\relax IEEE, 2023, pp. 1--5.

\bibitem{Lin_Gan_Han_2019}
\BIBentryALTinterwordspacing
J.~Lin, C.~Gan, and S.~Han, ``\BIBforeignlanguage{en-US}{Tsm: Temporal shift
  module for efficient video understanding},'' in
  \emph{\BIBforeignlanguage{en-US}{2019 IEEE/CVF International Conference on
  Computer Vision (ICCV)}}, Oct 2019. [Online]. Available:
  \url{http://dx.doi.org/10.1109/iccv.2019.00718}
\BIBentrySTDinterwordspacing

\bibitem{physnet}
Z.~Yu, W.~Peng, X.~Li, X.~Hong, and G.~Zhao, ``Remote heart rate measurement
  from highly compressed facial videos: An end-to-end deep learning solution
  with video enhancement,'' in \emph{2019 IEEE/CVF International Conference on
  Computer Vision (ICCV)}, 2019, pp. 151--160.

\bibitem{phsformer}
Z.~Yu, Y.~Shen, J.~Shi, H.~Zhao, P.~Torr, and G.~Zhao, ``Physformer: Facial
  video-based physiological measurement with temporal difference transformer,''
  in \emph{2022 IEEE/CVF Conference on Computer Vision and Pattern Recognition
  (CVPR)}, 2022, pp. 4176--4186.

\bibitem{pure}
R.~Stricker, S.~Müller, and H.-M. Gross, ``Non-contact video-based pulse rate
  measurement on a mobile service robot,'' in \emph{The 23rd IEEE International
  Symposium on Robot and Human Interactive Communication}, 2014, pp.
  1056--1062.

\bibitem{ubfc}
\BIBentryALTinterwordspacing
S.~Bobbia, R.~Macwan, Y.~Benezeth, A.~Mansouri, and J.~Dubois, ``Unsupervised
  skin tissue segmentation for remote photoplethysmography,'' \emph{Pattern
  Recognition Letters}, vol. 124, pp. 82--90, 2019, award Winning Papers from
  the 23rd International Conference on Pattern Recognition (ICPR). [Online].
  Available:
  \url{https://www.sciencedirect.com/science/article/pii/S0167865517303860}
\BIBentrySTDinterwordspacing

\bibitem{rppg_toolbox}
X.~Liu, G.~Narayanswamy, A.~Paruchuri, X.~Zhang, J.~Tang, Y.~Zhang,
  R.~Sengupta, S.~Patel, Y.~Wang, and D.~McDuff, ``rppg-toolbox: deep remote
  ppg toolbox,'' in \emph{Proceedings of the 37th International Conference on
  Neural Information Processing Systems}, ser. NIPS '23.\hskip 1em plus 0.5em
  minus 0.4em\relax Red Hook, NY, USA: Curran Associates Inc., 2023.

\bibitem{pure_ubfc_eval27}
Z.~Tu, H.~Talebi, H.~Zhang, F.~Yang, P.~Milanfar, A.~Bovik, and Y.~Li,
  ``Maxvit: Multi-axis vision transformer,'' in \emph{Computer Vision -- ECCV
  2022}, S.~Avidan, G.~Brostow, M.~Ciss{\'e}, G.~M. Farinella, and T.~Hassner,
  Eds.\hskip 1em plus 0.5em minus 0.4em\relax Cham: Springer Nature
  Switzerland, 2022, pp. 459--479.

\bibitem{pure_eva39}
\BIBentryALTinterwordspacing
S.~Bobbia, R.~Macwan, Y.~Benezeth, A.~Mansouri, and J.~Dubois, ``Unsupervised
  skin tissue segmentation for remote photoplethysmography,'' vol. 124, 2019,
  pp. 82--90, award Winning Papers from the 23rd International Conference on
  Pattern Recognition (ICPR). [Online]. Available:
  \url{https://www.sciencedirect.com/science/article/pii/S0167865517303860}
\BIBentrySTDinterwordspacing

\bibitem{ubfc_eva38}
R.~Stricker, S.~Müller, and H.-M. Gross, ``Non-contact video-based pulse rate
  measurement on a mobile service robot,'' in \emph{The 23rd IEEE International
  Symposium on Robot and Human Interactive Communication}, 2014, pp.
  1056--1062.

\end{thebibliography}

\vspace{12pt}
\color{red}
\end{document}